\documentclass{article}

\PassOptionsToPackage{numbers}{natbib}
\usepackage[dblblindworkshop, final]{neurips_2026}
\workshoptitle{RoboPAD 2026: Post-training Adaptation of Robot Foundation Models}

\usepackage[utf8]{inputenc}
\usepackage[T1]{fontenc}
\usepackage{hyperref}
\usepackage{url}
\usepackage{booktabs}
\usepackage{amsfonts}
\usepackage{amsmath}
\usepackage{amssymb}
\usepackage{nicefrac}
\usepackage{microtype}
\usepackage{xcolor}
\usepackage{graphicx}
\usepackage{siunitx}
\usepackage{wrapfig}
\usepackage{float}
\graphicspath{{figures/}}

\title{Behavioral Steering of Vision-Language-Action Models in the Low-Parameter Regime}

\author{%
  James Tcheng \qquad Feiyu Gavin Zhu\\
  Carnegie Mellon University\\
  \texttt{ztcheng@andrew.cmu.edu} \qquad \texttt{feiyuz@andrew.cmu.edu}
}

\begin{document}
\maketitle

\begin{abstract}
Reinforcement learning (RL) is increasingly used to adapt pretrained vision-language-action models (VLAs).
We investigate the problem of behavioral steering: adapting a VLA’s execution style to preferences such as carrying objects higher.
To investigate how little trainable capacity such steering requires and how much pretrained capability can be retained, we freeze an OpenVLA-OFT 7B policy and optimize TinyLoRA adapters with $8$ to $384$ trainable parameters using either policy-gradient RL (GRPO) or evolutionary search (CMA-ES).
With just $256$ trainable parameters, we raise carry height by $120$ mm on trained tasks at $92\%$ success and by $74$ mm on a held-out task at $80\%$ success.
By rescaling the learned update, we can tune the trade-off between style gains and capability retention without retraining.
Increasing adapter capacity generally improves this trade-off, retaining more pretrained capability at comparable style gains. The choice of optimizer also depends on capacity, with CMA-ES being competitive at low parameter counts, while GRPO offers better trade-offs as capacity grows.
These findings show that extremely low-dimensional updates can steer pretrained VLA behavior, with adapter capacity, optimizer choice, and update scaling offering practical control over adaptation and retention.

\end{abstract}
\section{Introduction}
\label{sec:intro}

The usefulness of a robot depends on both task completion and execution style. Policies that achieve the same goal may differ in speed, trajectory, or smoothness, with consequences for safety, efficiency, and user preference \citep{lasota2017survey,dragan2013legibility,story2022speed}.
Vision-language-action models (VLAs) achieve strong manipulation performance and generalize to new objects, scenes, and instructions \citep{openvla,brohan2023rt2,oneill2023openx}. Yet adapting their execution style while retaining their pretrained capabilities remains underexplored. 
As a case study, we steer carry height, following \citet{haon}, to measure how style gain and retention of those capabilities scale with the number of trainable
parameters.

Prior work adapts VLA execution style through training-free activation steering, which shifts hidden activations along directions associated with a concept \citep{actadd, caa} and can alter end-effector speed and carry height \citep{haon}. Since these directions are not optimized for the target behavior, neither the style change nor task success is guaranteed, and activation-level interventions in VLAs range from catastrophic to negligible across architectures and features \citep{grant2026}.
We find that the activation baseline based on \citet{haon}'s method does not consistently increase carry height and can disrupt task success, motivating a learned weight-space direction that directly targets the desired behavior.

Reinforcement learning (RL), increasingly used for VLA post-training \citep{rlinfvla, guo2025improvingvisionlanguageactionmodelonline, li2025simplevlarlscalingvlatraining}, is a natural way to learn such a direction. Many execution styles, such as carry height, speed, and smoothness, can be expressed as rewards without requiring style demonstrations. 
As the pretrained VLA already performs the task well, such adaptation may also need little capacity: in language models, RL with TinyLoRA improves math reasoning with only 13 trainable parameters, while supervised fine-tuning needs 100--1000$\times$ larger updates to match it \citep{tinylora}. We therefore study whether similarly small adapters suffice for VLA steering, and how adapter capacity and optimizer choice affect the trade-off between style gain and capability retention.

We study these questions empirically by steering a frozen OpenVLA-OFT 7B policy \citep{openvlaoft} to carry objects higher on LIBERO-Goal \citep{libero}. We train TinyLoRA adapters ranging from $8$ to $384$ parameters using a carry-height rollout reward subject to a task-success constraint, and compare two optimizers: GRPO \citep{grpo}, a policy-gradient method common in VLA post-training \citep{rlinfvla, li2025simplevlarlscalingvlatraining}, and CMA-ES \citep{cmaes}, a derivative-free evolution strategy increasingly explored for LLM post-training \citep{essa, esalternative, esgeometry}. 
We find that learned updates of a few hundred parameters
steer carry height where prompting and activation intervention do not, and rescaling them
recovers most of the held-out success they cost.
As capacity grows, GRPO gives a better style--retention trade-off than CMA-ES.

\paragraph{Contributions:}
\setlength{\leftmargini}{1.5em}
\begin{itemize}
  \item \textbf{Controlled study of VLA behavioral steering in the low-parameter regime.}
  We vary parameter count and optimizer independently, and evaluate every update on the full LIBERO-Goal suite
  alongside prompting and activation-intervention baselines (\S\ref{sec:steer}).

  \item \textbf{Matched-retention comparison of learned updates.}
  We use weight interpolation to trace each update's style--retention tradeoff curve, and compare
  updates at equal held-out success. This
  gives practitioners a way to tune the trade-off without retraining (\S\ref{sec:distance}, \S\ref{sec:dial}).

  \item \textbf{Analysis of how parameter count and optimizer shape steering.}
  For each optimizer, we identify how style at a fixed weight change scales with parameter count,
  separating the effect of added parameters from update size and training budget.
  We then compare the update directions GRPO and CMA-ES learn across parameter counts to understand why they differ (\S\ref{sec:optimizer_diff}).
\end{itemize}

\begin{figure}[t]
  \centering
  \includegraphics[width=\linewidth]{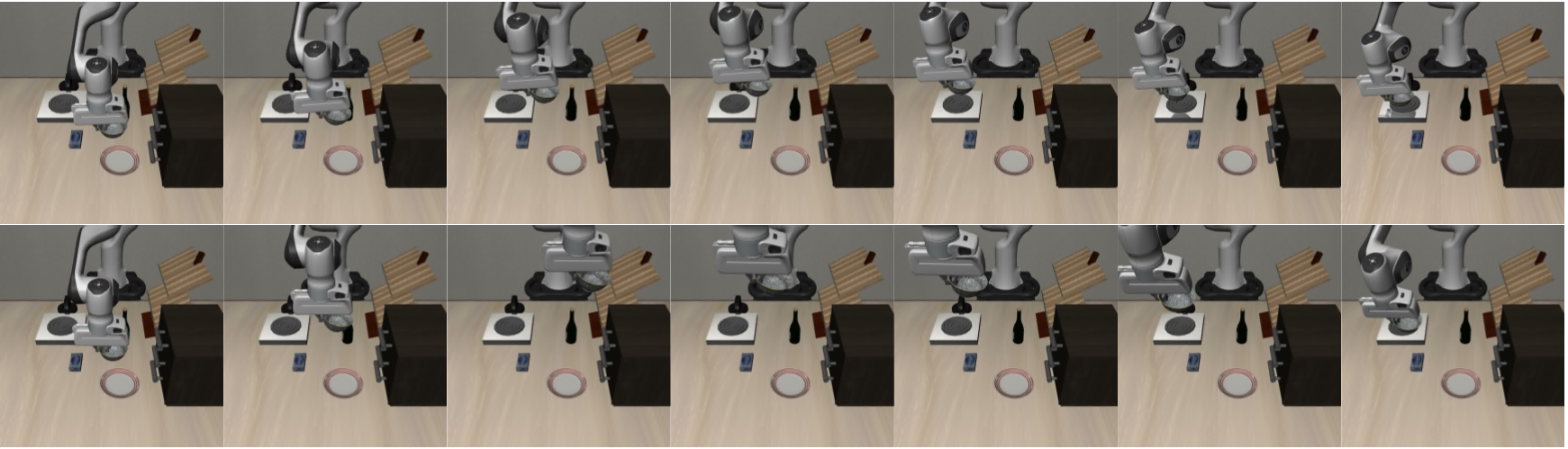}
  \caption{\textbf{Style transfer to a held-out task.} Our $64$-parameter update was trained on two other
LIBERO-Goal tasks, never on ``put the bowl on the stove'' (task~1). \textbf{Top:} the unsteered, frozen 7B
OpenVLA-OFT policy. \textbf{Bottom:} the same policy with the update learned with evolutionary strategies, which raises its peak carry height by
$+188$\,mm.}
  \label{fig:steered_vs_unsteered_teaser}
\end{figure}
\section{Related Work}
\label{sec:related}

\paragraph{Behavioral steering of VLAs.} 
Closest to our work, \citet{haon} steer VLAs without training by identifying FFN neurons whose value vectors project onto tokens expressing concepts such as fast, slow, and up, and overwriting their activations with fixed strength coefficients. They evaluate on OpenVLA in LIBERO simulation and on a physical UR5 robot controlled by $\pi_0$-FAST but do not report task success.
CTRL-STEER \citep{ctrlsteer} builds on this approach by adjusting the intervention strength during the rollout with a PID or RL controller. 
Both methods select neurons based on their association with concept words rather than the target behavior itself, and such interventions can also break task execution:
\citet{grant2026} finds that modifying even a single sparse-autoencoder feature can substantially reduce task success.
Beyond activation steering, TempoVLA \citep{tempovla} trains speed control into $\pi_{0.5}$ by fine-tuning on demonstrations re-timed by merging or splitting actions, but re-timing does not readily extend to other behavioral styles.
We instead learn weight updates directly from a behavioral rollout reward without demonstrations, comparing them against \citet{haon}'s activation steering and reporting how much held-out task success they retain.

\paragraph{Extremely parameter-efficient fine-tuning.} 
The intrinsic-dimensionality view of fine-tuning suggests that useful downstream adaptation of large pretrained models can lie in a much lower-dimensional
subspace \citep{intrinsic}.
LoRA \citep{lora} learns additive low-rank weight updates while keeping the pretrained weights frozen. 
Motivated by evidence that the pretrained weights' top singular directions provide an effective basis for fine-tuning, 
LoRA-XS \citep{loraxs} learns only a small matrix between fixed SVD factors. 
To further reduce trainable capacity, TinyLoRA \citep{tinylora} parameterizes this matrix through fixed random projections and parameter sharing, allowing as few as one trainable parameter. Trained with RL, it improves Qwen2.5-7B-Instruct's GSM8K accuracy from $76\%$ to $91\%$ using only 13 trainable parameters, which the authors suggest reflects a change in output style rather than new task knowledge.
We investigate whether similarly small adapters can steer a VLA's existing skills toward different execution styles.

\paragraph{Post-training with RL and ES.} 
Recent VLA systems use RL algorithms such as PPO \citep{ppo} and GRPO \citep{grpo} to fine-tune pretrained policies on task rewards, and report higher task success and better out-of-distribution performance than SFT-only baselines in simulation \citep{rlinfvla, li2025simplevlarlscalingvlatraining}. We similarly use GRPO, but optimize execution style under a task-success constraint.
Evolution strategies (ES) offer a backpropagation-free alternative, and have scaled to full-parameter fine-tuning of billion-parameter LLMs \citep{esatscale}. While ES can match GRPO's task accuracy, studies report larger weight changes, broader off-task drift \citep{esgeometry}, and greater forgetting of prior capabilities \citep{esforgetting}. We investigate whether the choice between ES and GRPO matters in the same way at our extremely small parameter counts, comparing both on style and on held-out task success.

\section{Method}
\label{sec:method}

We study stylistic adaptation of a frozen pretrained VLA policy by post-training only a small low-rank TinyLoRA adapter \citep{tinylora}. Following \citet{haon}, we focus on carry height: how high the robot holds an object while carrying it. To examine the effect of trainable capacity, we vary the adapter's parameter count $N_{\mathrm{train}}$ from a handful to a few hundred and compare optimization with CMA-ES and GRPO. Our experiments use the discrete-token OpenVLA-OFT policy from RLinf \citep{openvlaoft,rlinfvla}, based on the 7B OpenVLA model \citep{openvla, haon, ctrlsteer}. This checkpoint was trained on all 130 LIBERO tasks \citep{libero} through supervised fine-tuning on demonstrations followed by GRPO with task-success rewards.

\subsection{Low-rank adaptation with TinyLoRA}
\label{sec:param}
We adapt the frozen model with TinyLoRA \citep{tinylora}, an extremely parameter-efficient variant of
low-rank adaptation \citep{lora}.
Let the rank-$r$ truncated SVD of a
frozen weight be $W \approx U_r \Sigma_r V_r^\top$, where $U_r\in\mathbb{R}^{d_{\mathrm{out}}\times r}$ and $V_r\in\mathbb{R}^{d_{\mathrm{in}}\times r}$ hold the top-$r$ left and right singular vectors,
 and $\Sigma_r \in \mathbb{R}^{r\times r}$ is the diagonal matrix of the singular values. 
 LoRA-XS freezes this top-$r$ subspace and trains only a matrix
$R \in \mathbb{R}^{r\times r}$, updating $W$ by $U_r \Sigma_r R V_r^\top$. TinyLoRA allows us to drive
the trainable-parameter count even lower by expressing $R$ as a weighted sum of $u$ fixed random matrices $P_i$, with only the coefficients $\mathbf{v}\in\mathbb{R}^{u}$ learned:
\begin{equation}
  \Delta W = U_r\,\Sigma_r\Big(\textstyle\sum_{i=1}^{u} \mathbf{v_i} P_i\Big)V_r^\top
  \qquad P_i \sim \mathcal{N}\!\big(0,\tfrac{1}{ur}\big)
  \label{eq:tinylora}
\end{equation}
With $m$ adapted modules per layer over $n$ layers, and under tiled tying which groups every $n_{\mathrm{tie}}$ consecutive modules to share one vector $v$, the trainable-parameter count $N_{\mathrm{train}}$ is:
\begin{equation}
  N_{\mathrm{train}} = u\left\lceil \frac{nm}{n_{\mathrm{tie}}}\right\rceil
  \label{eq:count}
\end{equation}
We use $\theta\in\mathbb{R}^{N_{\mathrm{train}}}$ to denote the concatenation of all learned vectors $\mathbf{v}$ and use  $\pi_\theta$ for the base policy modified by the corresponding adapter update. Note that $\theta=\mathbf{0}$ recovers the frozen base policy. We measure how far an update moves the policy by \(\lVert \Delta W\rVert_F\), the aggregate Frobenius norm of the weight updates across all adapted modules. Since this distance is comparable across parameter counts, holding it fixed allows us to isolate the effect of update direction on style and retention.

\subsection{Carry height behavioral objective}
\label{sec:objective}
We steer how high the robot carries an object. Carry height is conceptually orthogonal to the task goal since it changes how a pick-and-place task is executed without changing what the robot is asked to accomplish. This makes it a natural behavioral axis across object-transport tasks.
For
a rollout $\tau\sim\pi_\theta$ on task $k$, let $(x_t,y_t,z_t)$ be the end-effector position at
timestep $t$, $\operatorname{success}(\tau)\in\{0,1\}$ indicate whether the task succeeds,
$\mathcal{C}(\tau)$ be the timesteps at which the robot is carrying an object, and $d_t = \lVert(x_{t+1},y_{t+1})-(x_t,y_t)\rVert_2$ be the horizontal displacement at step $t$. We define the style score as
\begin{equation}
    \operatorname{style}(\tau)=
    \frac{h_{\mathrm{carry}}(\tau)-h^{\mathrm{base}}_{k}}{\sigma_k}
    \qquad
    h_{\mathrm{carry}}(\tau)=\frac{\sum_{t\in \mathcal{C}(\tau)} d_t\,z_t}{\sum_{t\in \mathcal{C}(\tau)} d_t}
    \label{eq:style}
\end{equation}

where $h_{\mathrm{carry}}(\tau)$ averages end-effector height over the carry phase, weighting each timestep by its horizontal displacement. This weighting captures carry height during horizontal transport while limiting reward hacking through stationary holds, vertical-only motion, or deliberate slowing. We set
$h^{\mathrm{base}}_{k}$ to the frozen policy's median $h_{\mathrm{carry}}$ over successful task-$k$ rollouts,
and the spread $\sigma_k$ to the difference between the $95$th and $5$th percentiles of $h_{\mathrm{carry}}$
over those rollouts.

Our objective is to maximize the style score of successful trajectories while maintaining a task success rate above $p_{\min}$, so that carry height gains do not cost task completion:
\begin{equation}
    \max_{\theta}\;
    \mathbb{E}_{\tau\sim\pi_\theta}\!\left[\operatorname{style}(\tau) \mid \operatorname{success}(\tau)\right]
    \quad\text{s.t.}\quad
    \mathbb{E}_{\tau\sim\pi_\theta}\!\left[\operatorname{success}(\tau)\right]\ge p_{\min}
    \label{eq:objective}
\end{equation}

\subsection{Optimization with ES and GRPO}
\label{sec:opt}
We compare two optimizers from distinct families: CMA-ES \cite{cmaes}, a derivative-free evolution strategy, and GRPO \cite{grpo}, a policy-gradient method. Intuitively, CMA-ES sees only one scalar score per whole rollout, with no notion of which action
helped, while GRPO differentiates through the policy's log-probabilities to credit individual
actions within a rollout. Both optimizers search the trainable TinyLoRA adapter parameters $\theta\in\mathbb{R}^{N}$, initialized at the frozen base policy $\theta=\mathbf{0}$. Both optimizers target the constrained objective in Eq.~\eqref{eq:objective} through scalar surrogates that combine a style term with a penalty for falling below the minimum success rate $p_{\min}$.
To improve training stability and limit the influence of extreme style scores, we clip each rollout's style score using a threshold $c$:
\begin{equation}
\operatorname{style}_{\mathrm{clip}}(\tau)=\operatorname{clip}\bigl(\operatorname{style}(\tau),\,-c,\,c\bigr)
\label{eq:style-clip}
\end{equation}

\paragraph{Covariance Matrix Adaptation Evolution Strategy (CMA-ES).}
CMA-ES \citep{cmaes} is a derivative-free search that requires only a scalar score per candidate.
Each generation it draws a population
from $\mathcal{N}(m,C)$, then moves the mean $m$ toward the best candidates and
reshapes the covariance $C$, learning where to search and which
directions to spread along. 
We use the separable variant, which keeps $C$ diagonal and learns a step size for each coordinate independently so it adapts much faster than a full covariance at our parameter counts. 
Since adapter coordinates do not all move the weights equally, we
search in coordinates where one unit of distance is one unit of $\lVert\Delta W\rVert_F$.
A candidate's fitness is computed as its mean clipped style over its successful rollouts
$\mathcal{T}_\theta^{\mathrm{succ}}$, with a hinge penalty weighted by a fixed $\beta$, for how far its success rate $\hat p_\theta$ falls
below $p_{\min}$:
\begin{equation}
F(\theta)=\frac{1}{| \mathcal{T}_\theta^{\mathrm{succ}}|}\sum_{\tau\in \mathcal{T}_\theta^{\mathrm{succ}}}\operatorname{style}_{\mathrm{clip}}(\tau) - \beta\ \max(0, p_{\min}-\hat p_{\theta})
\end{equation}

\paragraph{Group-relative policy optimization (GRPO).}  Group-relative policy optimization \citep{grpo} is an on-policy policy-gradient
method, introduced for reinforcement-learning post-training of language models as a critic-free
variant of PPO \citep{ppo}. It samples multiple rollouts from the same initial state and normalizes their rewards within the group as the advantage, $A_i=(R(\tau_i)-\bar{R})/\sigma_{\mathrm{group}}$, replacing a learned value baseline. It then optimizes PPO's clipped objective over action tokens to favor above-average rollouts and discourage below-average ones.
 We omit the KL penalty to the reference policy. Following a Lagrangian approach to constrained RL \citep{stooke2020}, we reward successful rollouts by their clipped style and penalize failures
by $-\lambda$, adapting $\lambda$ to target a minimum success rate
$p_{\min}$:
\begin{equation}
    R(\tau)=\operatorname{success}(\tau)\,\operatorname{style}_{\mathrm{clip}}(\tau)
            -\big(1-\operatorname{success}(\tau)\big)\,\lambda
    \label{eq:grpofit}
\end{equation}

See Appendix~\ref{app:impl} for additional details.

\subsection{Weight interpolation}
\label{sec:distance}
In order to control how much of a trained adapter's style change we keep against how much
pretrained capability we retain, we interpolate between the pretrained and fine-tuned weights by
scaling $\theta\mapsto\alpha\theta$ from $\alpha=0$ (frozen base policy) to $\alpha=1$ (fully trained). This is an established technique for balancing adaptation and retention in both vision models
\citep{wiseft} and VLA policies \citep{vlamerge}. 
Since $\Delta W$ is linear in $\theta$ (Eq.~\ref{eq:tinylora}), this scales the weight update to $\alpha\Delta W$, tracing a success–style curve for each trained adapter without retraining. Along this curve, the distance from the frozen policy is $\alpha\lVert\Delta W\rVert_F$, letting us compare adapters at matched distances to separate update direction from magnitude.
\section{Experiments}
\label{sec:experiments}

\subsection{Experimental Setup}
\label{sec:setup}
\paragraph{Tasks.} Our experiments use LIBERO-Goal \citep{libero}, a suite of ten simulated tabletop manipulation tasks with different goals but the same objects and scene layout. We train on task 4 \textit{``put the bowl on top of the cabinet''} and task 8 \textit{``put the bowl on the plate''}, and hold-out the remaining eight tasks. 

\paragraph{Baselines.}
\label{sec:baselines}
We compare learned weight updates with two steering baselines:\begin{itemize}
  \item \textbf{Prompting}: prepending the style instruction ``high'' to the language instruction.
  \item \textbf{Activation intervention} \citep{haon}: a forward hook on the FFN's \texttt{down\_proj} layer clamps 20 keyword-selected ``up'' neurons to a fixed activation value of $10$ or $20$, overriding their activation during every forward pass.
\end{itemize}
\paragraph{Implementation details.}We adapt the FFN projections
$\{\texttt{gate\_proj},\texttt{up\_proj},\texttt{down\_proj}\}$ in all $32$ layers of the
Llama-2 language backbone of OpenVLA-OFT, fixing the rank at $r=2$. Under tiled tying, vectors are shared across the
three projections within each layer, so $u\in\{1,2,3,4,8,12\}$ gives
$N_{\mathrm{train}}\in\{32,64,96,128,256,384\}$ \eqref{eq:count}.
We also push the tying further at $u=1$, sharing one vector
across every $2$ and $4$ layers, which reaches $N_{\mathrm{train}}=16$ and $8$ respectively.
Actions are sampled from the policy's discrete head at
the recommended temperature $1.6$ \cite{rlinfvla}.
Both optimizers use $p_{\min}=0.9$ and clip style at $c=5$.
ES uses separable CMA-ES with mirrored sampling and a population of $24$, evaluating each candidate on $5$ rollouts per task. We run $128$ generations, increasing to $256$ for $N_{\mathrm{train}}\in \{256,384\}$.
GRPO
runs for $150$ iterations of $8$ groups of $8$ rollouts, taking one Adam step per iteration at
$\eta(N_{\mathrm{train}})=0.05\sqrt{96/N_{\mathrm{train}}}$, cosine-decayed to a $10\%$ floor ($200$ iterations at
$N\in\{256,384\}$). Remaining hyperparameters are in Appendix~\ref{app:impl}.

\paragraph{Evaluation.} Every evaluation is a full-suite LIBERO-Goal rollout on all
$10$ tasks $\times$ $50$ initial states. We also evaluate each update at several $\alpha$ scales (\S\ref{sec:distance}), including at matched $\lVert\Delta W\rVert_F$. Success on the eight held-out tasks measures how well pretrained capabilities are retained, while style and success on held-out task~1 ``put the bowl on the stove'' probes whether the steered behavior generalizes beyond the trained tasks.

\begin{table}[t]
  \centering
  \caption{Carry-height steering on LIBERO-Goal. \emph{Style} is the carry height style score \eqref{eq:style} in units of each task's own spread $\sigma_k$ ($\sigma_4{=}22.1$, $\sigma_8{=}22.0$, $\sigma_1{=}27.7$\,mm). This is averaged over the two trained tasks, task 4 and task 8. \emph{S.R.} is the task completion success rate. \emph{h.o.}\ is the mean over the eight held-out tasks. Left: the update as trained. Right: the same update scaled by $\alpha$ until held-out S.R.\ drops to $0.90$ and $0.75$, read from its $\alpha$ sweep.}
  \label{tab:main}
  \small
  \setlength{\tabcolsep}{3.5pt}
  \begin{tabular}{@{}lccccccccc@{}}
    \toprule
    & \multicolumn{5}{c}{as trained} & \multicolumn{2}{c}{h.o. S.R.\ $=0.90$} & \multicolumn{2}{c}{h.o. S.R.\ $=0.75$} \\
    \cmidrule(lr){2-6}\cmidrule(lr){7-8}\cmidrule(lr){9-10}
    & \multicolumn{2}{c}{trained tasks} & \multicolumn{2}{c}{task 1 (h.o.)} & h.o. & trained & task 1 & trained & task 1 \\
    \cmidrule(lr){2-3}\cmidrule(lr){4-5}\cmidrule(lr){6-6}\cmidrule(lr){7-7}\cmidrule(lr){8-8}\cmidrule(lr){9-9}\cmidrule(lr){10-10}
    & style & S.R. & style & S.R. & S.R. & style & style & style & style \\
    \midrule
      Unsteered & $0.00$ & $0.99$ & $0.00$ & $1.00$ & $0.96$ & --- & --- & --- & --- \\
      Prompt ``high'' & $+0.09$ & $1.00$ & $+0.04$ & $0.98$ & $0.96$ & --- & --- & --- & --- \\
      Activation, up, strength $10$ & $-0.31$ & $0.88$ & $+0.12$ & $0.98$ & $0.63$ & --- & --- & --- & --- \\
      Activation, up, strength $20$ & --- & $0.00$ & $-0.11$ & $0.16$ & $0.02$ & --- & --- & --- & --- \\
    \midrule
    \multicolumn{10}{@{}l}{\emph{ES}}\\
    \quad $N_{\mathrm{train}}{=}8$ & +0.23 & \textbf{0.97} & +0.35 & \textbf{1.00} & \textbf{0.91} & +0.24 & +0.35 & +0.36 & +0.37 \\
    \quad $N_{\mathrm{train}}{=}16$ & +1.20 & \underline{0.95} & +0.49 & \underline{0.98} & 0.38 & +0.39 & +0.24 & +0.62 & +0.37 \\
    \quad $N_{\mathrm{train}}{=}32$ & +1.70 & 0.89 & +0.70 & \textbf{1.00} & \underline{0.43} & +0.58 & \textbf{+0.54} & +1.00 & \underline{+0.75} \\
    \quad $N_{\mathrm{train}}{=}64$ & +3.44 & 0.89 & +2.17 & 0.66 & 0.11 & \underline{+0.62} & +0.43 & +1.09 & +0.71 \\
    \quad $N_{\mathrm{train}}{=}96$ & +4.83 & 0.88 & +1.28 & 0.96 & 0.33 & \textbf{+0.81} & \underline{+0.52} & +1.13 & +0.66 \\
    \quad $N_{\mathrm{train}}{=}128$ & +4.87 & \underline{0.95} & +2.00 & \underline{0.98} & 0.31 & +0.61 & +0.45 & \textbf{+1.37} & \textbf{+0.89} \\
    \quad $N_{\mathrm{train}}{=}256$ & \textbf{+5.78} & 0.92 & \textbf{+2.68} & 0.80 & 0.24 & +0.45 & +0.34 & \underline{+1.33} & +0.66 \\
    \quad $N_{\mathrm{train}}{=}384$ & \underline{+5.32} & 0.90 & \underline{+2.19} & 0.90 & 0.28 & +0.42 & +0.09 & +0.86 & +0.26 \\
    \addlinespace
    \multicolumn{10}{@{}l}{\emph{GRPO}}\\
    \quad $N_{\mathrm{train}}{=}8$ & +0.18 & \underline{0.95} & +0.08 & \textbf{1.00} & \underline{0.64} & +0.13 & +0.12 & +0.14 & +0.14 \\
    \quad $N_{\mathrm{train}}{=}16$ & +0.52 & \textbf{0.96} & +0.37 & \underline{0.98} & 0.43 & +0.33 & +0.46 & +0.42 & +0.54 \\
    \quad $N_{\mathrm{train}}{=}32$ & +1.21 & 0.81 & +0.62 & 0.94 & 0.39 & +0.68 & +0.32 & +0.82 & +0.27 \\
    \quad $N_{\mathrm{train}}{=}64$ & +1.69 & \textbf{0.96} & +0.83 & 0.96 & 0.51 & +0.61 & +0.30 & +0.95 & +0.51 \\
    \quad $N_{\mathrm{train}}{=}96$ & +1.51 & 0.93 & +0.54 & 0.96 & 0.42 & +0.45 & +0.24 & +0.91 & +0.26 \\
    \quad $N_{\mathrm{train}}{=}128$ & +1.78 & 0.89 & +0.98 & \textbf{1.00} & \textbf{0.76} & +1.16 & +0.66 & +1.78 & +1.08 \\
    \quad $N_{\mathrm{train}}{=}256$ & \underline{+2.93} & 0.93 & \underline{+1.96} & \underline{0.98} & 0.54 & \textbf{+1.65} & \underline{+0.98} & \textbf{+2.33} & \underline{+1.43} \\
    \quad $N_{\mathrm{train}}{=}384$ & \textbf{+3.16} & 0.88 & \textbf{+2.61} & 0.88 & 0.37 & \underline{+1.26} & \textbf{+1.01} & \underline{+1.97} & \textbf{+1.48} \\
    \bottomrule
  \end{tabular}
\end{table}

\subsection{Few trainable parameters are enough to steer carry height}
\label{sec:steer}

Learned updates produce substantially larger style gains than the training-free baselines
(Table~\ref{tab:main}). On tasks 4 and 8, prompting yields a style gain of only $0.09$.
Activation intervention at strength $10$ lowers style by $0.31$ and reducing held-out
success to $0.63$, while at strength $20$ the policy collapses with almost no
successful rollouts. By contrast, at full strength, a $256$-parameter ES update increases trained-task style by $5.78$ ($119.9$ mm on task 4 and $134.6$ mm on task 8) at $0.92$ success,
while a GRPO update of the same size increases it by $2.93$ ($85.0$ mm on task 4 and $44.2$ mm on task 8) at $0.93$ success.
Both updates also increase carry height on the untrained
bowl-to-stove task: ES raises style by $2.68$ ($74.3$ mm) at $0.80$ success, and GRPO
by $1.96$ ($54.5$ mm) at $0.98$ success. Qualitative examples appear in
Appendix~\ref{app:qualitative}.

\begin{figure}[!t]
  \centering
  \includegraphics[width=\linewidth]{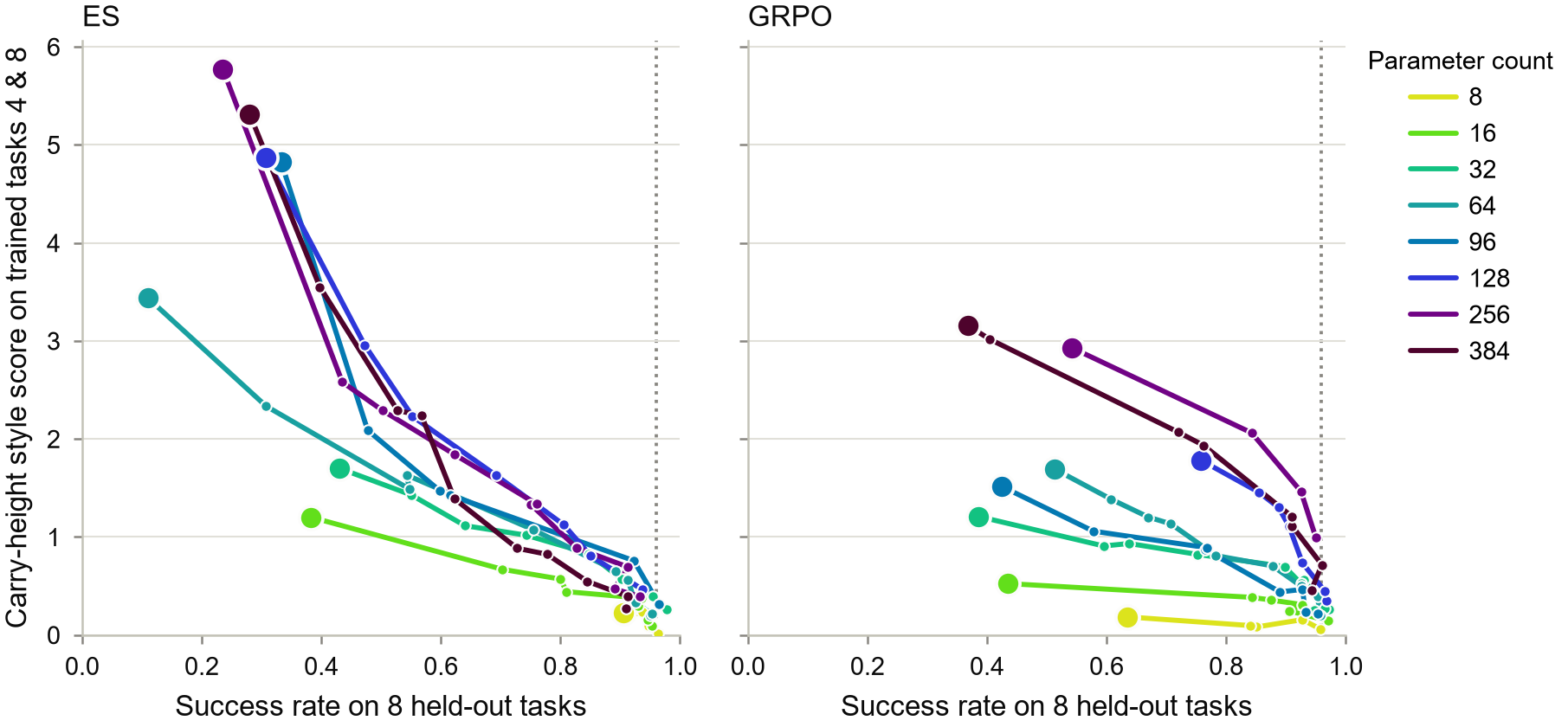}
  \caption{\textbf{Sweeping $\alpha$ traces the style--retention trade-off.} Carry-height style on the trained tasks
against success on the eight held-out tasks, ES left and GRPO right. Each curve is a single trained
update rescaled by $\alpha$ (\S\ref{sec:distance}). The large marker is the trained update, the small markers are its
rescalings, and color is the parameter count. Dotted line: unsteered held-out success at $0.96$.}
  \label{fig:transfer}
\end{figure}

\subsection{Trade-off between style gain and retention of pretrained capabilities}
\label{sec:dial}
Much of the lost held-out success can be recovered while preserving meaningful style gains, without retraining. As trained, the highest-style ES and GRPO updates reduce success on the eight held-out tasks from $0.96$ for the base policy to $0.24$ and $0.37$, respectively.
Rescaling each update by $\alpha$ (\S\ref{sec:distance}) traces a style--retention curve between the base policy and the update as trained  (Figure~\ref{fig:transfer}), allowing this trade-off to be tuned.
For example, GRPO at $256$ parameters raises carry height by $65$\,mm at held-out success of
$0.54$. Scaled to $\alpha=0.60$, it still raises carry height by $32$\,mm, half its full effect, while held-out success returns to $0.93$.

Increasing the number of trainable parameters generally improves the style--retention trade-off, yielding higher style at comparable held-out success (Figure~\ref{fig:transfer}). These gains level off at different parameter budgets for the two optimizers. For ES, $96$--$128$ parameters offer the strongest trade-offs over much of the moderate-to-high retention range. Larger budgets reach higher maximum style, yet often achieve less style at comparable retention. GRPO benefits from scaling to $256$ parameters, whose curve offers higher style than $384$ over most of their shared retention range despite a slightly lower maximum style score. The budget that maximizes steering at full strength therefore need not offer the best trade-off after rescaling.

\begin{figure}[H]
   \centering
  \includegraphics[width=0.48\linewidth]{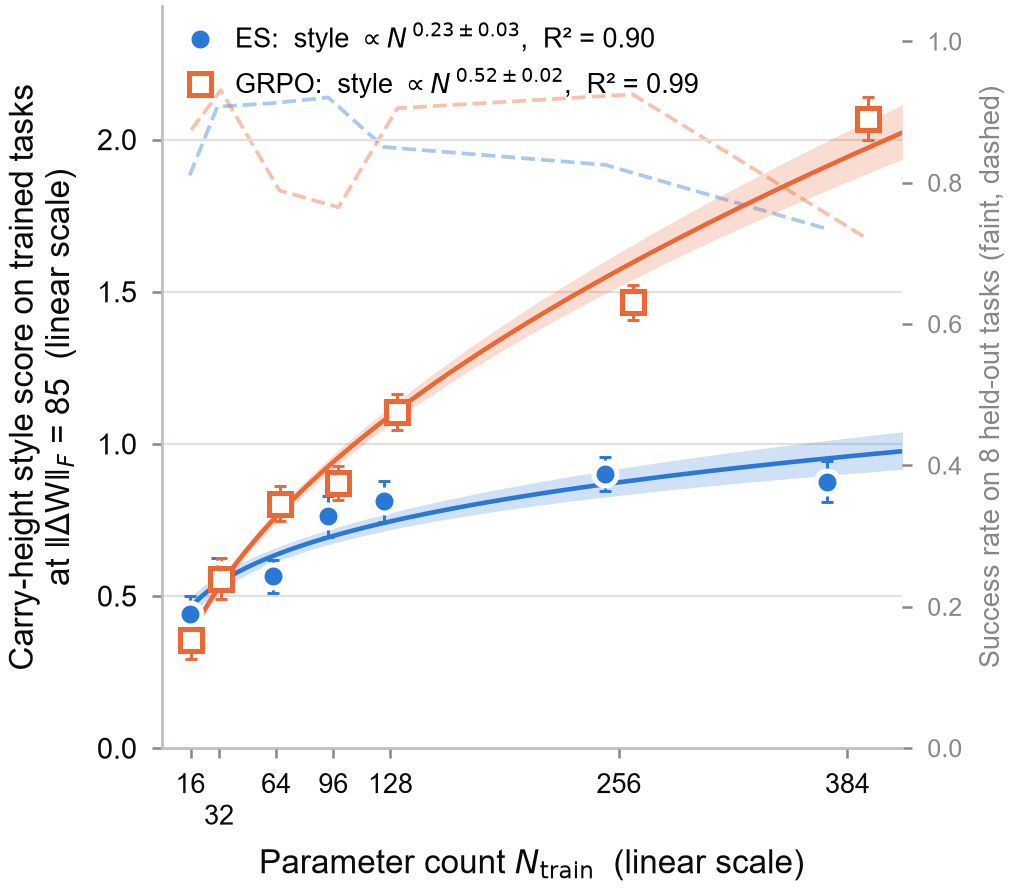}
  \hfill
  \includegraphics[width=0.48\linewidth]{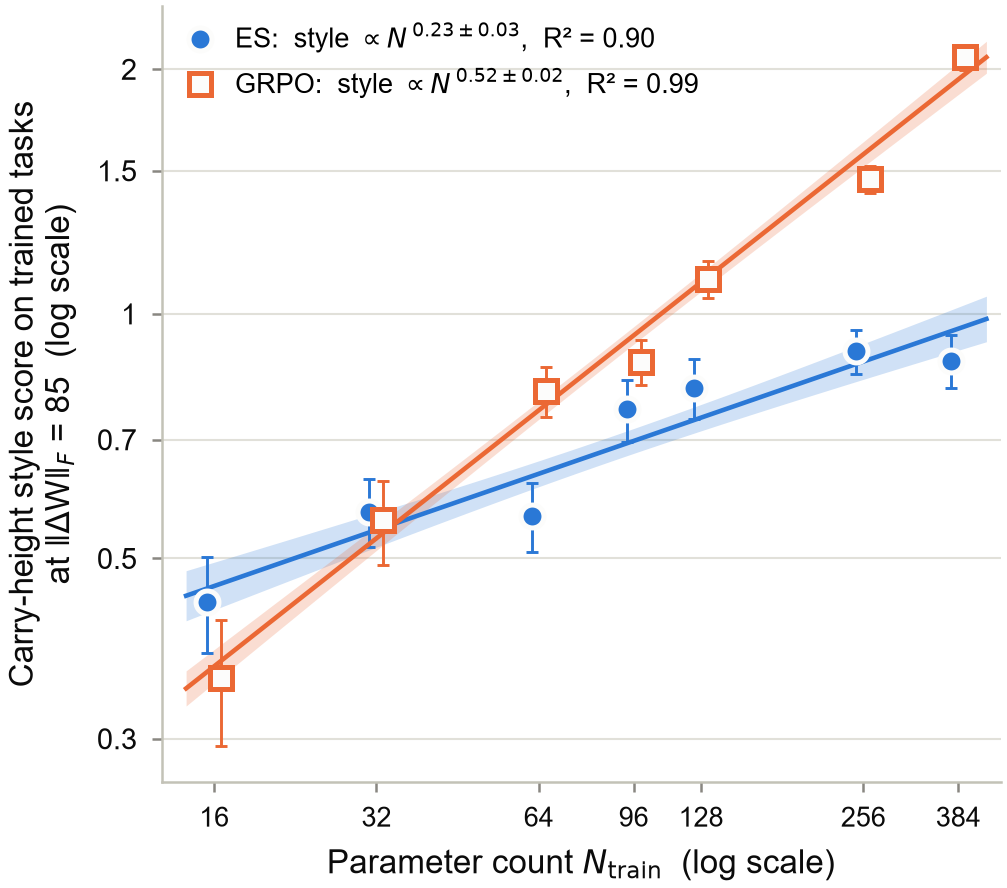}
  \caption{\textbf{At a fixed weight change, style grows sub-linearly with parameter count for both optimizers, while GRPO's style grows faster with parameter count than ES's.}
  Carry-height style score on the trained tasks at a fixed weight change,
  against parameter count $N_{\mathrm{train}}$. Style grows sub-linearly in both, at $N^{0.52 \pm 0.02}$ ($R^2 = 0.99$) for GRPO and
  $\propto N^{0.23 \pm 0.03}$ ($R^2 = 0.90$) for ES. Faint dashed lines (left panel, right axis) show success on the eight held-out tasks at the same
  weight change. The 8-parameter cells are absent because their trained updates
  stop short of $85$ units.}
  \label{fig:scaling}
\end{figure}

\subsection{Comparison between ES and GRPO}
\label{sec:optimizer_diff}

GRPO achieves a better style--retention trade-off than ES at larger parameter counts. 
At held-out success $0.90$ and $0.75$, its rescaled updates achieve higher style than ES at $128$, $256$,
and $384$ parameters (Table~\ref{tab:main}). This advantage is reflected in
the curve shapes: from $128$ parameters up, GRPO's curves are concave,
gaining style steeply while held-out success remains high, whereas ES's
curves from $96$ parameters up are convex, with substantial style gains
emerging only after larger losses in held-out success.

GRPO also benefits more strongly from additional trainable capacity than ES.
Since ES produces larger full-strength updates than GRPO, averaging $186$ units against $138$ for
$N\geq32$, we compare the optimizers at a fixed weight change to isolate the
effect of added parameters from update size.  At $\lVert\Delta W\rVert_F=85$, style scales with parameter count as $N^{0.52\pm0.02}$ for GRPO, compared with $N^{0.23\pm0.03}$ for ES, with similar held-out success (Figure~\ref{fig:scaling}). 
Additional capacity therefore yields more style for GRPO at the same weight change and comparable task success.
GRPO's update directions are also more consistent across parameter counts. Its updates have a mean pairwise cosine similarity of $0.41$ versus $0.17$ for ES, with higher similarity for every corresponding pair (Table~\ref{tab:direction_alignment}). Since style at a fixed weight change depends only on direction, these results indicate that GRPO
keeps finding useful directions as capacity grows, whereas ES's directions vary from count to count.

We attribute GRPO's stronger scaling to how the two optimizers identify useful directions. GRPO uses policy gradients to guide updates, whereas CMA-ES relies on rankings of sampled parameter perturbations, an estimate that becomes less informative as the  dimension of the search space grows. This interpretation is consistent with prior findings that ES can accumulate movement along directions weakly related to its objective \citep{esgeometry}.

\begin{table}[t!]
  \centering
  \small
  \setlength{\tabcolsep}{4pt}
  \caption{\textbf{GRPO's updates stay aligned across parameter counts while ES's become less aligned as the
count grows.} 
Each entry reports the mean cosine similarity between a given optimizer's weight update $\Delta W$ at parameter count $N_{\mathrm{train}}$ and that same optimizer's updates across the remaining five parameter counts.
$1$ means the same direction, $0$
unrelated.}
  \label{tab:direction_alignment}
  \begin{tabular}{lrrrrrr}
    \toprule
    $N_{\mathrm{train}}$  & 32 & 64 & 96 & 128 & 256 & 384 \\
    \midrule
    GRPO & 0.36 & 0.41 & 0.37 & 0.49 & 0.46 & 0.38 \\
    ES   & 0.23 & 0.20 & 0.21 & 0.17 & 0.11 & 0.10 \\
    \bottomrule
  \end{tabular}
\end{table}

\section{Limitations \& Future Work}
\label{sec:limitations}
\paragraph{Limitations.} We evaluate only carry-height steering on LIBERO-Goal in simulation, without real-robot validation. Each configuration uses one training seed, so intervals capture evaluation variability only. Hyperparameters were tuned lightly, and some differences between ES and GRPO may stem from unequal training budgets (ES uses $2.4$--$4.8\times$ more episodes) and differing surrogates of the same constrained objective. Style transfers to held-out bowl tasks but weakly to wine-bottle and multi-step tasks. Our metric measures only transport height and is undefined without a carry phase. Extending to new styles requires careful metric design and validation to mitigate reward hacking.

\paragraph{Future work.} Extending evaluation to other tasks, VLA architectures, and real robots would test the generality of our findings, including whether steering extends to diffusion and flow-matching policies. Beyond carry height, learning and composing updates for speed, clearance, and precision could enable combined styles such as ``slow and low''. Training on more diverse tasks may also improve transfer across objects and task sequences. Finally, multi-seed comparisons with matched budgets and identical objectives would clarify the relative strengths of ES and GRPO and test whether the observed crossover with parameter count persists.

\section{Conclusion}
\label{sec:conclusion}
We studied carry-height steering of a frozen OpenVLA-OFT 7B policy on LIBERO-Goal, using TinyLoRA adapters optimized with GRPO or CMA-ES under a task-success constraint. 
 A few dozen trainable parameters produced substantial carry-height gains that transferred to a held-out task, while prompting and activation interventions yielded little or no improvement. 
These findings suggest that meaningful behavioral adaptation in VLAs can have a very low effective dimensionality.
Rescaling learned updates provides a practical way to balance style gains and retention of pretrained capabilities without retraining. Additional capacity generally improves this trade-off, with GRPO making better use of additional trainable capacity than CMA-ES. Very few parameters can induce substantial behavioral change, while a larger adaptation space helps achieve that change with less disruption to pretrained capabilities.

\section*{Acknowledgment}

This work was in part supported by the Institute of Information \& Communications
Technology Planning \& Evaluation(IITP) grant funded by the Korea
government(MSIT) (No.RS-2025-25463654).
This research used computing resources on Bridges-2 at the Pittsburgh Supercomputing Center, supported by NSF award OAC-1928147, through ACCESS allocation CIS260236. The Advanced Cyberinfrastructure Coordination Ecosystem: Services \& Support (ACCESS) program receives funding under NSF awards 2138259, 2138286, 2138307, 2137603, and 2138296.

Feiyu Zhu is supported by the Softbank Group-Arm PhD Fellowship.
This research is part of the Keio-CMU partnership.

{\small
\bibliographystyle{unsrtnat}
\bibliography{references}
}
\appendix
\newpage
\section{Implementation details}
\label{app:impl}

\paragraph{Policy and rollouts.} We use the RLinf-OpenVLAOFT-LIBERO-130 checkpoint \cite{rlinfvla} with its discrete
action-token head, sampling actions at
temperature $1.6$ from a single center-cropped camera view, with no proprioceptive input. Each policy call
emits a chunk of $8$ actions, and episodes run for at most $300$ steps after a $10$-step settle. Both
optimizers train on tasks $4$ and $8$ of LIBERO-Goal using all $50$ initial states of each, and we
report the final iterate of every run (the final CMA mean for ES, the last iteration for GRPO) with no
validation-based selection. Each run uses a single seed (ES $7$, GRPO $0$).

\paragraph{Objective.} The success requirement is $p_{\min}=0.9$ and the style clip is $c=5$. Per-task spread is measured as $\sigma_4{=}22.1$, $\sigma_8{=}22.0$, $\sigma_1{=}27.7$\,mm. We detect the carry phase $\mathcal{C}(\tau)$ as the set of steps on which the gripper is commanded closed and the object has begun moving, so the reported gains are carries rather than
early-gripper-close artifacts. 

\paragraph{CMA-ES.} Separable CMA-ES with mirrored sampling, population $24$, penalty $\beta=10$, run for $128$
generations ($256$ for $N\in\{256,384\}$). Each candidate is scored on $5$ rollouts per task. We
search in coordinates in which Euclidean distance equals weight-space distance by construction,
obtained by applying the inverse matrix square root of the exact block-diagonal Gram form that
relates each adapter coordinate to $\lVert\Delta W\rVert_F$. In these coordinates the initial step
size is $\sigma_0 = 78.4/\sqrt{N}$. The final CMA mean
is used as the trained update.

\paragraph{GRPO.} Each iteration collects $8$ groups of $8$ rollouts ($64$ episodes), $4$ groups per
trained task. Each group starts from one initial state.
We use the PPO clipped surrogate (ratio bounds $[0.8,\,1.28]$) with no KL or entropy term,
micro-batches of $8$ policy calls, gradient clipping at norm $1$, and one Adam step per iteration. A cosine learning-rate
schedule from $\eta_0=0.05\sqrt{96/N}$ down to $10\%$ of $\eta_0$. This is chosen so that update steps have similar length in parameter space across counts. Each task's penalty $\lambda_k$ starts at $0$ and
is updated after every step as
$\lambda_k\leftarrow\operatorname{clip}\big(\lambda_k+0.5\,(p_{\min}-\hat p_k),\,0,\,10\big)$, where
$\hat p_k$ is the task's success rate in that iteration.

\paragraph{Compute.} Each run used two NVIDIA RTX 4090 or two NVIDIA L40S GPUs for training and evaluation. An average ES run takes $4.6$\,h while an average GRPO run takes $7.8$\,h.

\newpage

\section{Qualitative Behavior}
\label{app:qualitative}

\begin{figure}[H]
  \centering
  \includegraphics[width=\linewidth]{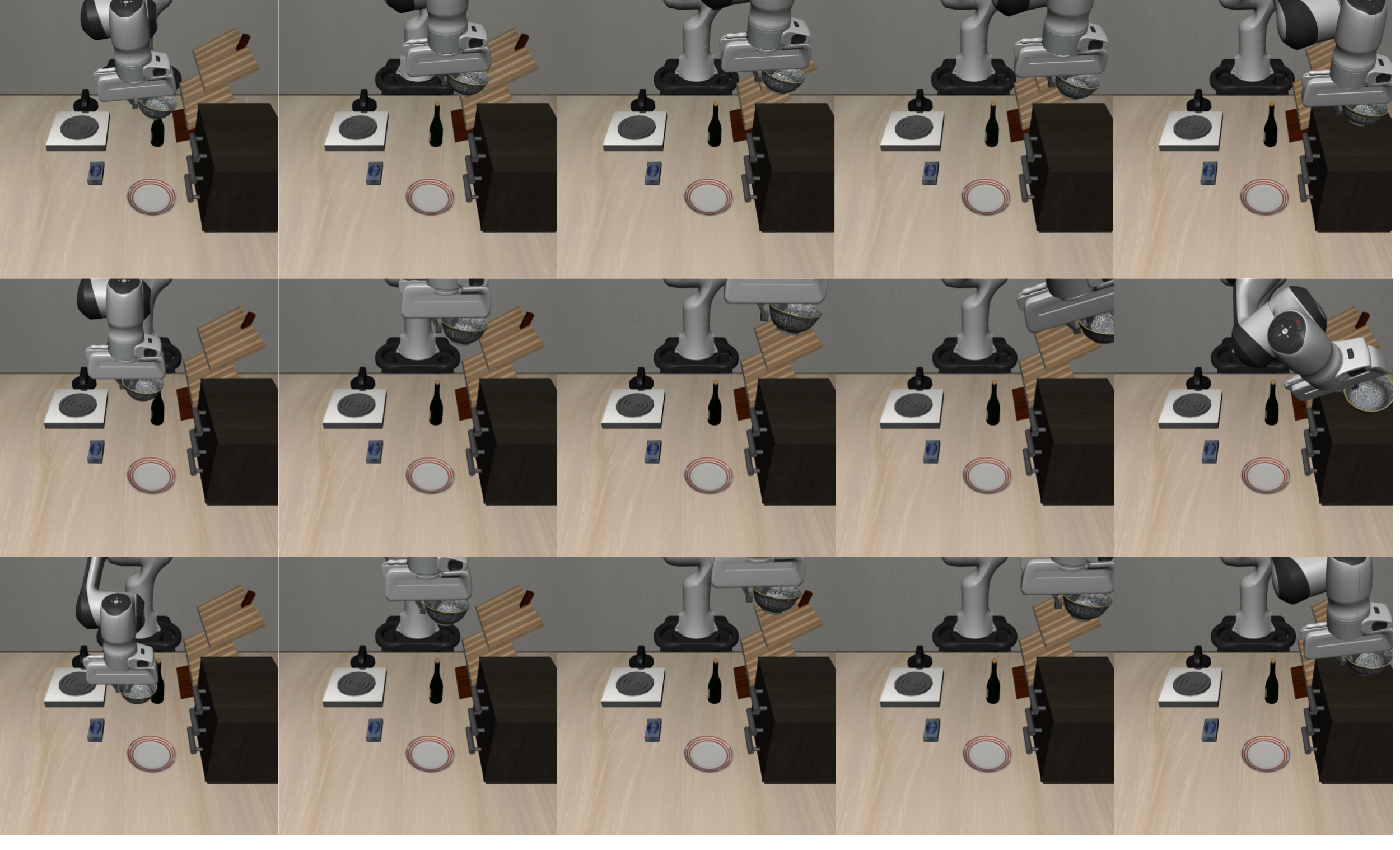}
  \caption{\textbf{Learned updates raise carry height on task~4, ``put the bowl on top of the
  cabinet''.} One episode per policy from the same initial state (state~$6$). \emph{Top:} unsteered
  policy. \emph{Middle:} ES with $N_{\mathrm{train}}{=}256$, carrying the bowl $108.3$\,mm higher at peak.
  \emph{Bottom:} GRPO with $N_{\mathrm{train}}{=}256$, $95.9$\,mm higher at peak. Both updates are shown as
  trained ($\alpha{=}1$).}

$ $\\
\includegraphics[width=\linewidth]{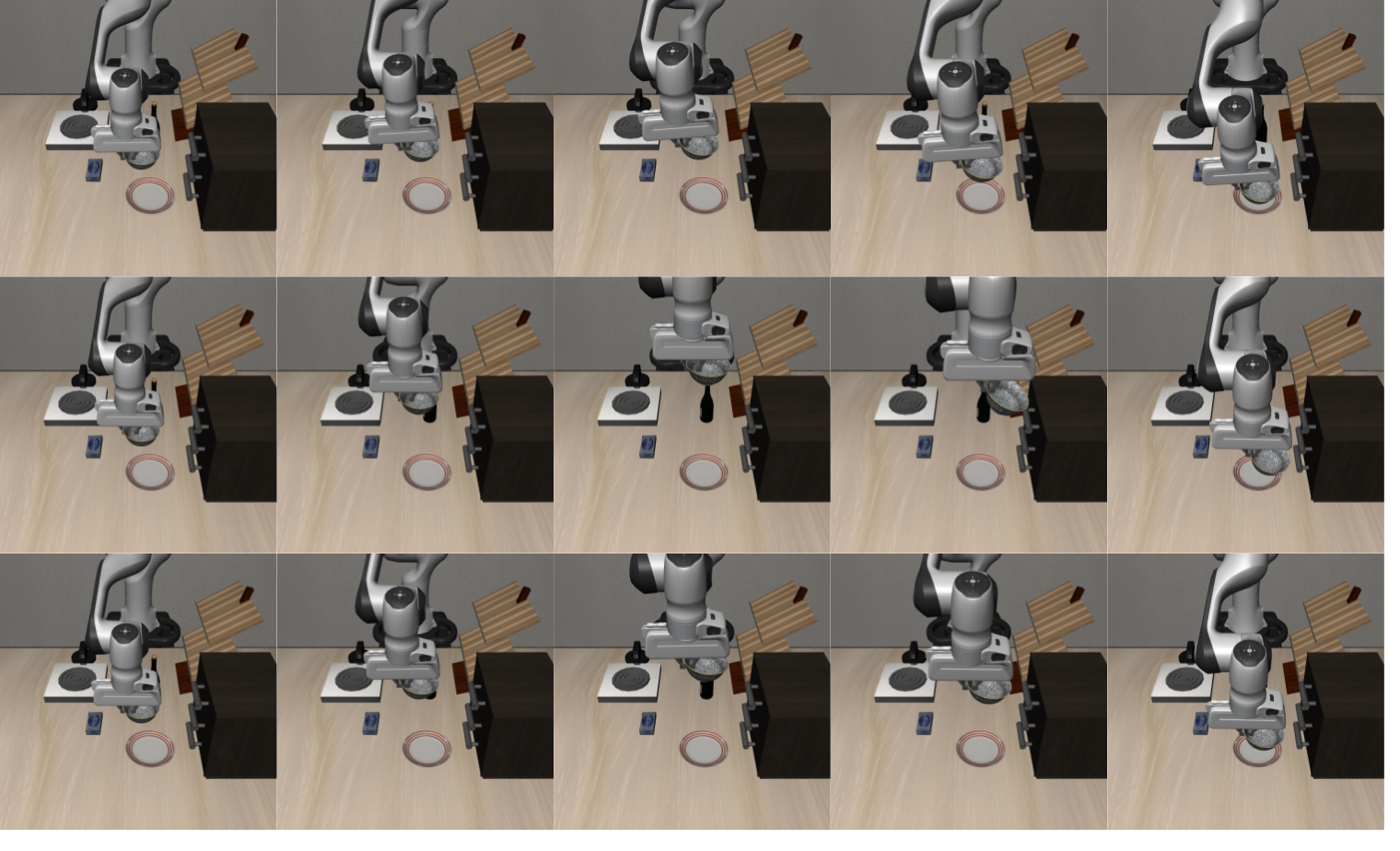}
  \caption{\textbf{Learned updates raise carry height on task~8, ``put the bowl on the plate''.} One episode per policy from the same initial state (state~$14$). \emph{Top:} unsteered
  policy. \emph{Middle:} ES with $N_{\mathrm{train}}{=}256$, carrying the bowl $155.2$\,mm higher at peak.
  \emph{Bottom:} GRPO with $N_{\mathrm{train}}{=}256$, $108.3$\,mm higher at peak. Both updates are shown as
  trained ($\alpha{=}1$).}

  \label{fig:rollouts}
\end{figure}

\end{document}